\relax
\documentclass[letterpaper]{article} 
\usepackage{aaai19}  
\usepackage{times}  
\usepackage{helvet}  
\usepackage{courier}  
\usepackage{url}  
\usepackage{graphicx}  

\begin{document}
%
\title{Agent-Based Adaptive Level Generation for Dynamic Difficulty Adjustment in Angry Birds}

\author{Matthew Stephenson,\textsuperscript{1,2}
Jochen Renz,\textsuperscript{1}\\
\textsuperscript{1}{Research School of Computer Science, Australian National University, Canberra, A.C.T. 0200, Australia}\\
\textsuperscript{2}{Department of Data Science and Knowledge Engineering (DKE), Maastricht University, Netherlands}}

\maketitle

\begin{abstract}
This paper presents an adaptive level generation algorithm for the physics-based puzzle game Angry Birds. The proposed algorithm is based on a pre-existing level generator for this game, but where the difficulty of the generated levels can be adjusted based on the player's performance. This allows for the creation of personalised levels tailored specifically to the player's own abilities. The effectiveness of our proposed method is evaluated using several agents with differing strategies and AI techniques.
By using these agents as models / representations of real human player's characteristics, we can optimise level properties efficiently over a large number of generations.
As a secondary investigation, we also demonstrate that by combining the performance of several agents together it is possible to generate levels that are especially challenging for certain players but not others.
\end{abstract}

\section{Introduction}
Procedural level generation (PLG), where levels for a game are created automatically without the need for human designers, is a key area of investigation for video game research \cite{ori1,ori2}. PLG can be extremely useful for increasing a game's length and replayability, as it allows a large number of levels to be created in a relatively short time. It is also possible to tailor the generated levels towards specific user's playstyles, known as adaptive level generation, which allows for a unique and personalised gameplay experience \cite{ori4}. Dynamic difficulty adjustment (DDA) is a form of adaptive level generation, where the difficulty of generated levels is modified to better suit the player's current skill based on their performance \cite{ori5}. This is accomplished by modifying certain generator parameters that control different level features, so that the end result is more likely to achieve the desired amount of challenge for the player.

This paper presents an adaptive level generator for the physics-based puzzle game Angry Birds. This game has been used substantially in AI research over the past few years, primarily for developing agents and level generators, as the game's environment presents more realistic physical constraints compared to most traditional video games. Successfully generating levels for Angry Birds that are equally as challenging as human-designed levels is a difficult task, but will likely be necessary for Angry Birds agents to improve beyond their current capabilities.
Previous level generation methods for Angry Birds used either a heuristic calculation based on level properties or the performance of several agents to help set the difficulty of a level.
However, as different players often possess varying levels of ability, many people would likely find these levels too hard or easy to solve. This is also a problem for training and evaluating agents, as levels where most agents either can or cannot solve them yield very little discriminatory information \cite{minecont}.
We therefore suggest an agent-based adaptive generation method for dynamic difficulty adjustment, where the generator adjusts the difficulty of its levels depending on the player's performance.
This method can also be used to generate levels that are difficult for one player whilst being easy for another, exploiting the player's own strengths or weaknesses.

The remainder of this paper is organised as follows. Section 2 describes the large amount of background and related work, both for Angry Birds and adaptive level generation in general. Section 3 presents our proposed adaptive generation method. Section 4 describes our conducted experiments and results. Sections 5 discusses what these results could mean for both human players and agents, Section 6 concludes this work and outlines future possibilities.

\section{Background}

\subsection{Adaptive Level Generation}
While most games that contain some form of PLG typically use generic generation techniques that are not influenced by the player's behaviour, adaptive level generation, also referred to as experience-driven, personalised or player-centred level generation, takes the player's behaviour into account \cite{pcgbook}.
Examples of game or level characteristics that could be adjusted for specific players include qualities such as difficulty, engagement, frustration, enjoyment, complexity, learning potential, etc. These properties are indirectly controlled by adjusting certain parameters of the generator. Different players will likely behave or perform differently even when playing the same game. If an accurate model of the player can be determined, then this can be used to tailor the properties of the generated content towards their individual preferences.

Constructing a model of the player is a difficult and imprecise task, but is essential for adaptive level generation to be effective. Methods for determining player behaviour include analysing their performance across several ``test'' levels, or using a questionnaire for measuring more intangible qualities. This information can then be used to directly evaluate generated content in the future, allowing us to estimate whether it will be suitable for the player. Another approach, and the one that we will be using in this paper, is to use AI agents to estimate the quality of levels (i.e. agent / simulation-based evaluation functions). Using a collection of agents as representations of different playing styles or abilities allows us to generate levels that are suited to a particular player, or a collection of levels that require several different techniques to solve them.

Examples of genre's where adaptive level generation has been used to improve the player experience include board games \cite{other8}, racing games \cite{other9}, action-RPG \cite{other11}, rogue-like \cite{other10}, tower defence \cite{other2}, and platformers \cite{other6,other7}.

\subsubsection{Dynamic Difficulty Adjustment}
Dynamic difficulty adjustment (DDA) is often considered to be one of the simplest and most common forms of game adaption, where the difficulty of a game increases or decreases if the player is performing too well or poorly respectively \cite{other5}. Because a player's performance in a game can typically be evaluated without the need for questionnaires or overly complex estimations, DDA can usually be implemented in most games without significant issue. Nearly all games feature some form of increasing difficulty as the skill of the player increases, but this element is often lost or overly simplified with most PLG approaches. However, just because estimating the difficulty of a game for a specific player is relatively simple compared to other more complex behavioural characteristics, this certainly doesn't make the task trivial. The difficulty of a game can often be multi-dimensional in nature, where the same level could be considered hard or easy for various different reasons \cite{other4}. Players can often have unbalanced skill sets, where they are adept at overcoming certain tasks or challenges more than others. One player may be very good at forward planning, another at making precise actions, another with fast response times, and so on. A successful DDA system should therefore be able to adjust the difficulty of its generated levels in many different ways, and also detect which of these most influence the player's performance.

\subsubsection{Agent-Based Evaluation}
One approach for evaluating generated content is to utilise AI agents with different strategies to play through the generated levels \cite{other5,other7,other9}. By selecting the agent that best models the player's abilities, we can then use this agent as a player surrogate in the adaptive level generation process.
This approach has several benefits. First, agents can often be used to play levels much faster than a normal person, allowing us to evaluate a larger number of levels in a much shorter time (requiring volunteers to playtest hundreds of levels is not very practical). Agents can also typically give more accurate estimations of certain level properties (especially difficulty) than by just analysing the level's features. Human players are also likely to improve the longer they play, making repeated performances inconsistent between different experiments. The downside of this method is that it naturally requires a large and diverse range of agents to already exist, which Angry Birds thankfully has (agents described in more detail later).

\subsection{Angry Birds}
Angry Birds is a popular physics-based puzzle game where the player's objective for each level is to kill all pigs using a set number of birds. A typical Angry Birds level like the one shown in Figure 1, requires the player to shoot the birds they have from a slingshot at structures made of blocks that are protecting the pigs. All objects within the level obey simplified physics principles defined by the game's engine. Blocks can come in several different shapes and materials, and birds can also be one of several different types (all with differing properties). Pigs and blocks can be killed / destroyed by hitting them with either a bird or another object. Points are awarded to the player once the level is solved (all pigs in the level have been killed) based on the number of birds remaining and the total amount of damage caused. The source code for the official Angry Birds game is not currently available, so a modified version of the Unity-based clone known as Science Birds, originally created by Lucas Ferreira \cite{lucas1}, was used instead. 

\begin{figure}
    \centering
  \includegraphics[width=1.0\linewidth]{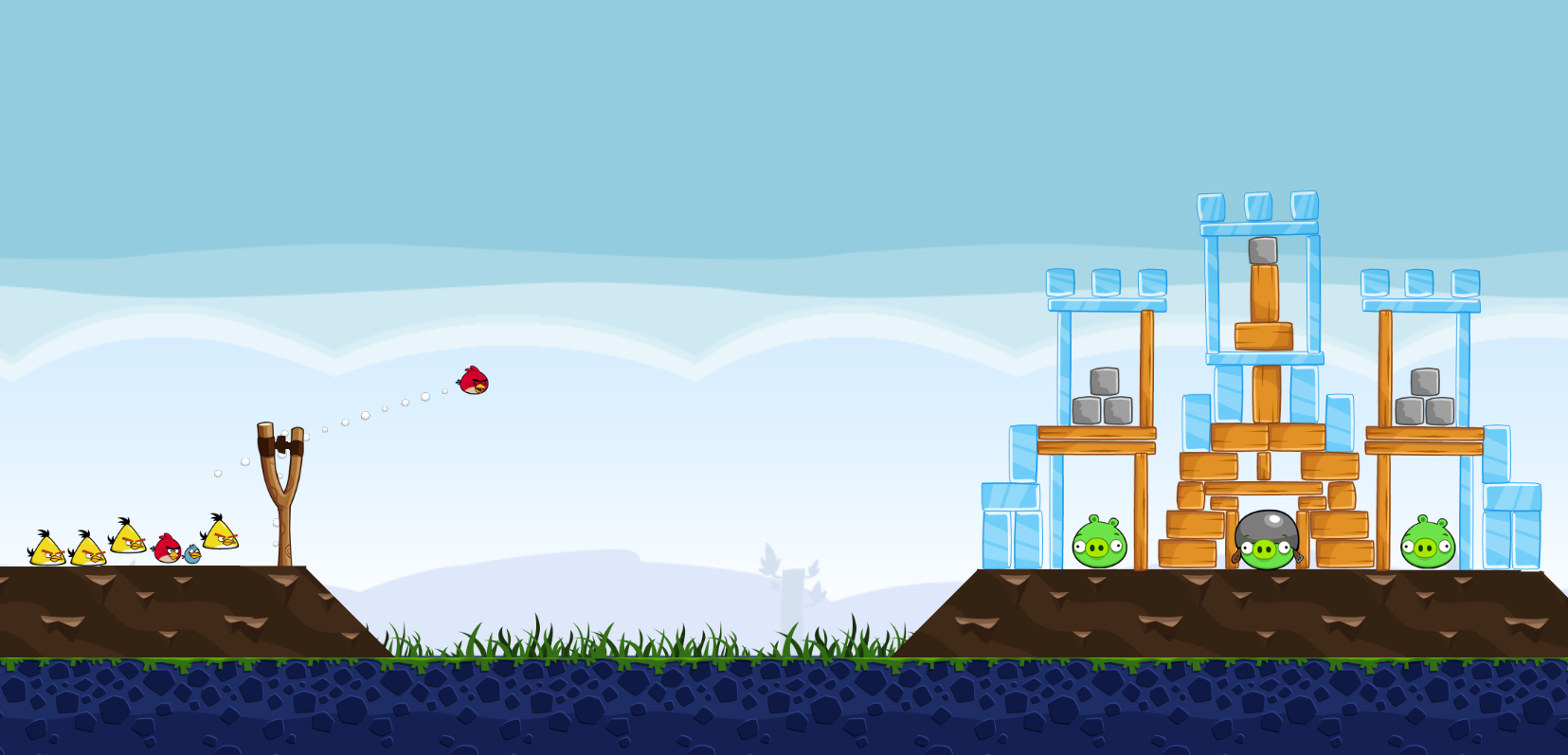}
  \caption{Screenshot of a level from the Angry Birds game.}
\end{figure}

\subsubsection{Level Generation}
Several level generators have been presented for Angry Birds in recent years, some of which have attempted to adapt the generated content based on the player's experience.
Previous work by \cite{kaidan1,kaidan2} attempted to measure the difficulty of Angry Birds levels based on their features, and take this into account during the generation process. The generator they present is based on the same genetic algorithm described in \cite{lucas1}, but where the fitness function for evaluating generated levels has been modified to take the desired difficulty of the level into account. The first approach simply used the number of pigs within a level as a measure of difficulty \cite{kaidan1}. The desired number of pigs for each level would then be adjusted over multiple generations, based on the number of pigs that the player was able to kill in previous levels.
An alternative measure of difficulty was proposed in a subsequent paper \cite{kaidan2}, which attempted to estimate the difficulty of a level based on its overall impact factor. This was calculated based on the ERA relations between objects within the level. In both these prior cases, the fitness function for the generator rewards levels with an estimated difficulty closest to the desired amount, which in turn is based on the player's performance for previous levels. However, neither of these approaches use agents to evaluate levels and their estimations of difficulty are based solely on a level's features, which are controlled by only a small number of generator parameters.

Instead of changing the difficulty, some prior generators investigated other level aspects that might influence the player's experience.
The Tanager generator evaluated the immersion and design quality of its generated levels using an on-line user study \cite{lucas2}. This was in the form of a questionnaire which asked users to rate both automatically and manually created levels in terms of their enjoyment, engagement and challenge.
The Funny quotes generator creates levels based on words or quotes for three levels of difficulty \cite{funny1}. Certain generator parameters were manually configured for each difficulty category, based on the results of a user study into the average solve and retry rates of players across different levels. 
A follow up investigation using a similar version of this generator, modified future levels based on chat comments made by players inside the game \cite{funny2}. Words within these comments were used as input parameters when generating future levels (i.e. the generator adjusts the levels it creates based on what the players type to each other).

The adaptive generator presented in this paper is based on the Iratus Aves generator described in \cite{mine5,mine4,mine3}, which was also the winner of the 2017 and 2018 Angry Birds level generation competitions \cite{new88}. The output of this generator can be partially controlled by changing the values of different input parameters (i.e. the generator's parameter set). This search-based generator previously used a direct fitness function approach to modify generator parameters over several generations based on desirable level properties. Instead of evaluating a generated level based solely on its observable features, we implement a new agent-based fitness function which uses the performance of several Angry Birds agents. Angry Birds agents have been utilised for a small number of previous generators \cite{mine5,lucas2}, but only to check if a generated level is solvable. This paper uses agents to evaluate and evolve the generated levels in a deeper and more meaningful way.

\subsubsection{Agents}
A wide variety of Angry Birds agents have been developed over the past six years for the AIBirds competition \cite{abrenz1,abrenz2,abrenz3}. These agents employ a range of different AI techniques, including qualitative reasoning \cite{cite3}, internal simulation analysis \cite{cite5,cite4}, logic programming \cite{cite10}, heuristics \cite{cite11}, Bayesian inferences \cite{cite7,cite6}, and structural analysis \cite{cite8}. In this paper we selected four different agents to assist with evaluating generated levels. These were the Naive, Datalab, SeaBirds and Eagle's Wing agents. The Naive agent is the simplest agent available, making it a perfect model of a novice player. The remaining three agents (referred to as ``skilled agents'') are some of the best performing agents currently available \cite{mine1,mine2}, although they are still well below that of a normal human, with each agent having their own strengths and weaknesses \cite{mine6}. Further details about the specific strategies and AI techniques used by each of these agents can be found in \cite{mine1}.

\section{Methodology}
To reiterate our proposed method using previous terminology, we present an agent-based evaluation function for level generation, which allows for dynamic difficulty adjustment in Angry Birds and other similar physics-based puzzle games. To achieve this, we need both a way to evaluate the player's performance and a way to update the level generator's output based on this performance. By using a search-based generation approach, we can evolve an initial population of parameter sets for our level generator over many generations using a fitness function \cite{other4,other6}.

A general overview of the adaptive level generation process is as follows:
\begin{enumerate}
\item Measure the performance of the player and all available agents on a randomly generated collection of levels, and select the agent that best models the player (e.g. lowest root-mean-square error).
\item Randomly create an initial population of parameter sets (individuals) for our level generator.
\item Generate a level for each individual in the population and record each agent's performance on these levels.
\item Use these agent performance distributions to calculate a fitness value for each individual in the population. 
\item Evolve this population using a genetic algorithm (selection, crossover, mutation, elitism, etc.) based on each individual's fitness value (i.e. create a new generation).
\item Stop once a desired number of generations has been reached; otherwise repeat from step 3.
\end{enumerate}

By following this process, the average fitness of the parameter sets (and levels generated using them) within our population should increase over multiple generations.

Note that step 1 of this process essentially selects an agent to act as a representation of our player in all subsequent steps, and is therefore unnecessary if the player is already an agent (i.e. only needed for human players).

\subsection{Adaptive Level Generator}
As previously mentioned, our proposed adaption method is based on the same Angry Birds level generator described in \cite{mine5,mine4,mine3}. This generator previously used a fitness function to evaluate and update the probability of selecting different block shapes based on certain features of the generated levels. For our adaption method to be successful, we need to be able to control more level properties than just the frequency of block shapes. We therefore extended the number of input parameters that affect the generated levels, see left column of Table 1. Apart from the increased number of adjustable input parameters, the level generation algorithm itself was not changed.

\begin{table}[t]
\begin{center}
\begin{tabular}{|p{5.0cm}|p{2.4cm}|}
    \hline
    \textbf{Generator Input Parameter} & \textbf{Value Range}\\ \hline
    	Number of pigs & 1 - 15 (integer)\\ \hline
	Number of birds & 1 - 8 (integer)\\ \hline
	Number of ground structures & 1 - 5 (integer)\\ \hline
	Number of platform structures & 0 - 4 (integer)\\ \hline
	Maximum number of TNT & 0 - 4 (integer)\\ \hline
	Weights for each bird type (x5) & 0.0 - 1.0 (float)\\ \hline
	Weights for each material (x3) & 0.0 - 1.0 (float)\\ \hline
	Weights for each block shape (x13) & 0.0 - 1.0 (float)\\ \hline
  \end{tabular}
\caption{Input parameters for our adaptive level generator and their possible value range (minimum - maximum).}
\end{center}
\end{table}

The first four parameters define the number of pigs, birds and structures (both on the ground and platforms) within the generated level. The fifth parameter determines the maximum number of TNT boxes that the level can contain (could potentially be less that this value depending on the available space). The last three parameters are lists of values that define weightings for each bird type (five options), material (three options) or block shape (thirteen options). Unlike the previous parameters, these weight inputs do not directly define specific level features, but instead influence the probability of selecting their respective elements (i.e. if the weight value of one block shape is twice that of another, then that block shape has twice the chance of being selected during level generation). While all weight inputs are float values between zero and one, integer inputs are limited to within a fixed value range, see right column of Table 1.
Each parameter set within our population contains values for each of these generator input parameters (genome length of 26).

\subsection{Difficulty Estimation}
Whilst prior methods for estimating the difficulty of an Angry Birds level relied solely on its observable features, we instead propose a more accurate approach based on agent performance.
This allows us to not only better estimate the difficulty of levels overall, but also means that the same level can be given multiple difficulty scores based on different player's abilities. Angry Birds has two basic measures of success. The first is simply solving each level and the second is achieving a large score for each level, with the score for a level being awarded after it is solved. This score element to solving levels allows for an additional degree of depth when comparing different agents. Perhaps one agent solves a level less often than another agent, but typically achieves a higher score when it does. We therefore proposed two possible difficulty measures ($D_{solve}$ and $D_{score}$) of a level ($L$) for an agent ($A_i$), see Equations 1 and 2.

\begin{small}
\begin{equation}
{D_{solve}(A_i, L) = 1 - \frac{\#TimesSolved(A_i, L)} {\#Attempts(A_i, L)}}
\end{equation}
\end{small}
\begin{small}
\begin{equation}
{D_{score}(A_i, L) = 1 - \frac{AverageScore(A_i, L)} {MaximumScore(L)}}
\end{equation}
\end{small}

Both $D_{solve}$ and $D_{score}$ can be any value between zero and one (normalised).
$MaximumScore(L)$ is defined as the theoretical score that could be achieved if all pigs and blocks within $L$ were destroyed using only the first bird.

Essentially, $D_{solve}$ uses the agent's solve-rate for a level as the measure of difficulty, whilst $D_{score}$ uses the score-rate. Deciding which difficulty measure to use depends on the desired property of the generated levels.

\subsection{Fitness Function}
Now that we can estimate the difficulty of a level for a specific agent, we can define fitness functions that use this to evaluate the parameter sets within our population. The fitness value for each parameter set is based on the difficulty measures of our agents for a level generated using it. Many different fitness functions could be defined that each represent a desired performance distribution of our agent(s), but we will only focus on two in this paper.

The first function defines the fitness of a level in terms of the probability that our agent is able to solve the level each time they attempt it, see Equation 3, where $A_{s}$ is the specific agent that the generated levels are being adapted for, $D_{solve}$ is the observed solve-rate, and $D_{target}$ is the target / desired solve-rate. 

\begin{small}
\begin{equation}
{Fitness_p(A_{s}, L) = 1 - abs(D_{solve}(A_{s}, L) - D_{target})}
\end{equation}
\end{small}

This allows us to define the desired difficulty of a generated level for a specific agent as a percentage (i.e. if we want an agent to solve each generated level 50\% of the times it attempts it, then we simply set $D_{target}$ to 0.5). $D_{score}$ could also be used as our difficulty measure for this function instead of $D_{solve}$, but trying to define a desired score-rate for a level as a fraction of the total score possible is conceptually harder to understand than simply the desired solve-rate.

The second fitness function is more complex and utilises several different agents. Instead of adapting our generated levels to a fixed solve-rate for a specific agent, it is also possible to adapt our generated levels to be especially hard for our chosen agent when compared to the performance of other agents, see Equation 4, where $A$ is the set of all available agents.

\begin{small}
\begin{equation}
{Fitness_m(A_s, L) = D_{score}(A_s, L) - {\min\limits_{A_n \in A}({D_{score}(A_n, L)}})}
\end{equation}
\end{small}

Using this fitness function will favour levels that our specific agent performs poorly in, but where other agents perform better. Essentially, adapting the generated levels using this fitness function will focus on our specific agent's weaknesses, while using the inverse of this function will generate levels that focus on its strengths. Using $D_{score}$ as our difficulty measure rather than $D_{solve}$ allows us to still compare the performances of different agents, even when their solve-rates are very similar (i.e. using score-rate gives a more precise measure of performance then solve-rate).

To summarise, $Fitness_{p}$ gives a higher value to levels that are closer to the desired solve-rate for a specific agent, whilst $Fitness_{m}$ gives a higher value to levels that a specific agent finds relatively difficult compared to other agents.

\subsection{Genetic Algorithm}
Once a fitness value for each parameter set in our current population has been calculated, a genetic algorithm is used to evolve the population and create the next generation. Individuals are selected from the current population using stochastic universal sampling \cite{bak1}. This selection technique reduces the risk of individuals with a large fitness value being overrepresented in the next population (i.e. gives individuals with a lower fitness a greater chance of being chosen). This is desirable, as the uncontrollable stochastic elements of our generator and agents means that the fitness value for each parameter set is likely to be only a rough estimate of its actual fitness. An elitism scheme was also used to select a percentage number of individuals in each generation with the highest fitness value, and include these in the next generation unchanged. Uniform crossover and mutation genetic operators were then used to create the new generation (offspring) from the previously selected individuals of the current generation (parents). Mutations for each parameter set value must be within the possible minimum and maximum range for that parameter, as described in Table 1.

\section{Experiments and Results}
Two experiments were conducted using our proposed adaptive level generation algorithm, for each of the fitness functions previously described. The first experiment investigated our adaptive generator's ability to create levels with a desired solve-rate ($Fitness_p$) for both a novice player (Naive agent) and an expert player (hyper-agent created from the skilled agents). The second experiment investigated whether our adaptive generator could successfully create levels that skilled agents performed better on than the Naive agent ($Fitness_m$ with the Naive agent as $A_s$), essentially generating levels that exploited the Naive agent's limitations.

The specific values used for our genetic and level evolution algorithms are as follows. Parameter sets were adapted over 30 generations, with a population size of 50 individuals. Elitism was set at a rate of 8\%, crossover probability at 25\%, and mutation probability at 15\%. Each agent was given a three minute time limit to play each generated level on a heavily sped up version of the modified Science Birds game. While the number of attempts each agent was able to complete for each generated level can fluctuate depending on the agent's design and the level's features, each agent took roughly 2-3 seconds between making each shot. This means that even if a generated level contained the maximum number of birds (eight), each agent would still get at least six or seven attempts to solve it.

Due to the high variance in our generator's output and each agent's performance, the individual results of our experiments were highly stochastic. We therefore repeated each experiment ten times to reduce any inaccuracy issues, with the results displayed in the following sections being the averaged results over all ten repeated experiments. Please also note that even though our experiments were run over 30 generations the graphs displaying our results only show the first 20 generations, as the average fitness of our generated levels never increased significantly past this point.

\begin{figure}
    \centering
  \includegraphics[width=1.0\linewidth]{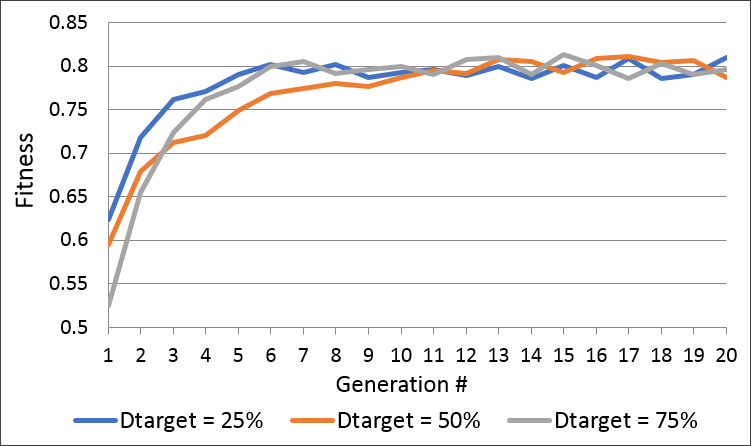}
  \caption{Average $Fitness_p$ value of each generation for the Naive agent.}
\end{figure}

\begin{figure}
    \centering
  \includegraphics[width=1.0\linewidth]{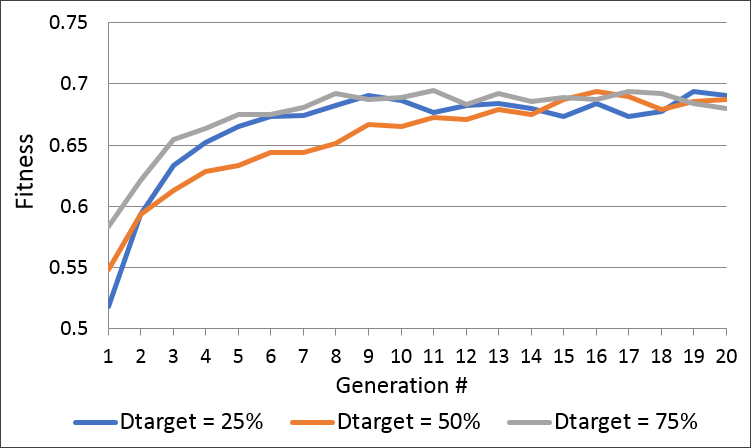}
  \caption{Average $Fitness_p$ value of each generation for the skilled hyper-agent.}
\end{figure}

\subsection{Percentage Solvability}
This experiment investigated the effectiveness of our $Fitness_p$ function for the Naive agent and different $D_{target}$ values.
We also performed the same analysis for a hyper-agent that selects from our three skilled agents (Datalab, SeaBirds and Eagle's Wing), based on the same score prediction models described in \cite{mine2}. Three $D_{target}$ values were tested, 25\%, 50\% and 75\%.
The average $Fitness_p$ values over all parameter sets in each generation, for both the Naive agent and the skilled hyper-agent, are shown in Figures 2 and 3 respectively.
From these results we can see that the average fitness of the generated levels for all agents and $D_{target}$ values increased over the generations tested. However, the rate at which the fitness increased and the optimal fitness value that could be reasonably achieved, appears to be different for each agent and $D_{target}$ pairing.

When using a $D_{target}$ value of 50\%, our adaptive generator took longer to reach a high fitness value compared to the other $D_{target}$ values. This was likely because levels that could never be solved and levels that could always be solved had an equal $Fitness_p$ value. Due to the large number of highly variable parameters that can influence the difficulty of a generated level, it would be very easy for our adaptive generator to only produce levels that would probably be impossible to solve, by simply making the number of pigs and structures very high whilst also making the number of birds very small. The opposite can also be done to generate levels that are incredibly easy to solve. Both of these types of levels are also very likely to occur in the randomly generated initial population. As a result of these factors, it is easier for our adaptive generator to create levels for the lower or higher $D_{target}$ values in the earlier generations, as it can initially focus on simply creating either impossibly hard or ridiculously easy levels respectively. Using a $D_{target}$ value of 50\% treats both these cases as equally desirable, meaning that our adaptive generator must find a suitable balance between the two. This naturally takes more time to accomplish, but over a large number of generations the average fitness of the generated levels eventually equals that of the other $D_{target}$ values.

Comparing individual $D_{target}$ values, it also appears that adapting generated levels for the skilled hyper-agent took longer to reach a high fitness value when compared to the Naive agent. The maximum fitness value that could be achieved also appeared to be less for the skilled hyper-agent, only around 0.69 compared to the Naive agent's maximum fitness of around 0.81.
This was likely due to the skilled hyper-agent have more strategies and behaviours that must be ``learned'' by our level adaption algorithm (i.e. combinations of multiple level properties probably required to construct levels of a suitable difficulty).

\subsection{Relative Solvability}
This experiment investigated the effectiveness of our $Fitness_m$ function for evolving levels that the Naive agent performed poorly on relative to the performance of more skilled agents. Generated levels should not only be hard for the Naive agent (which could easily be achieved using the $Fitness_p$ function and setting $D_{target}$ to a very small value) but should also be easier for the skilled agents to solve with a larger score. If successful, this would essentially create levels that require a certain degree of skill to perform well on, an idea that is often represented within traditional human-designed levels.
The average $Fitness_m$ values over all parameter sets in each generation are shown in Figure 4.
Please note that as $Fitness_m$ is calculated using the $D_{score}$ measure, which is based on the theoretical maximum score that could possibly be obtained for a level and is often significantly higher than any realistically achievable score, the $Fitness_m$ values for levels are significantly lower than the previous $Fitness_p$ values.

\begin{figure}
    \centering
  \includegraphics[width=1.0\linewidth]{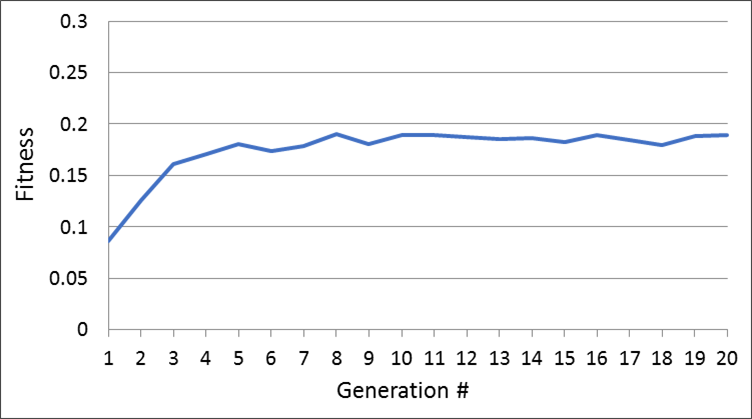}
  \caption{Average $Fitness_m$ value of each generation for the Naive agent.}
\end{figure}

From these results we can see that the average fitness of the generated levels increased slightly over the generations tested.
This result is promising, as it means that it is possible to generate levels that favour certain agents over others. 
By manually comparing the parameter sets of the generated levels with the highest $Fitness_m$ values it would seem that, apart from simply being harder overall, levels that the Naive agent struggled the most with compared to the skilled agents contained more TNT boxes and bird types with difficult to use abilities (yellow, blue and white birds). This makes sense, as our Naive agent doesn't directly target TNT and doesn't vary the tap time for activating bird's abilities (unlike the skilled agents). This observation is also backed up by a previous investigation into deceptive Angry Birds level design \cite{mine6}.

An example of a level that was generated using an evolved parameter set based on our $Fitness_m$ function is shown in Figure 5. This level had a (relatively) high fitness value of 0.62, indicating that at least one of the skilled agents was able to significantly outperform the Naive agent.

\begin{figure}
    \centering
  \includegraphics[width=1.0\linewidth]{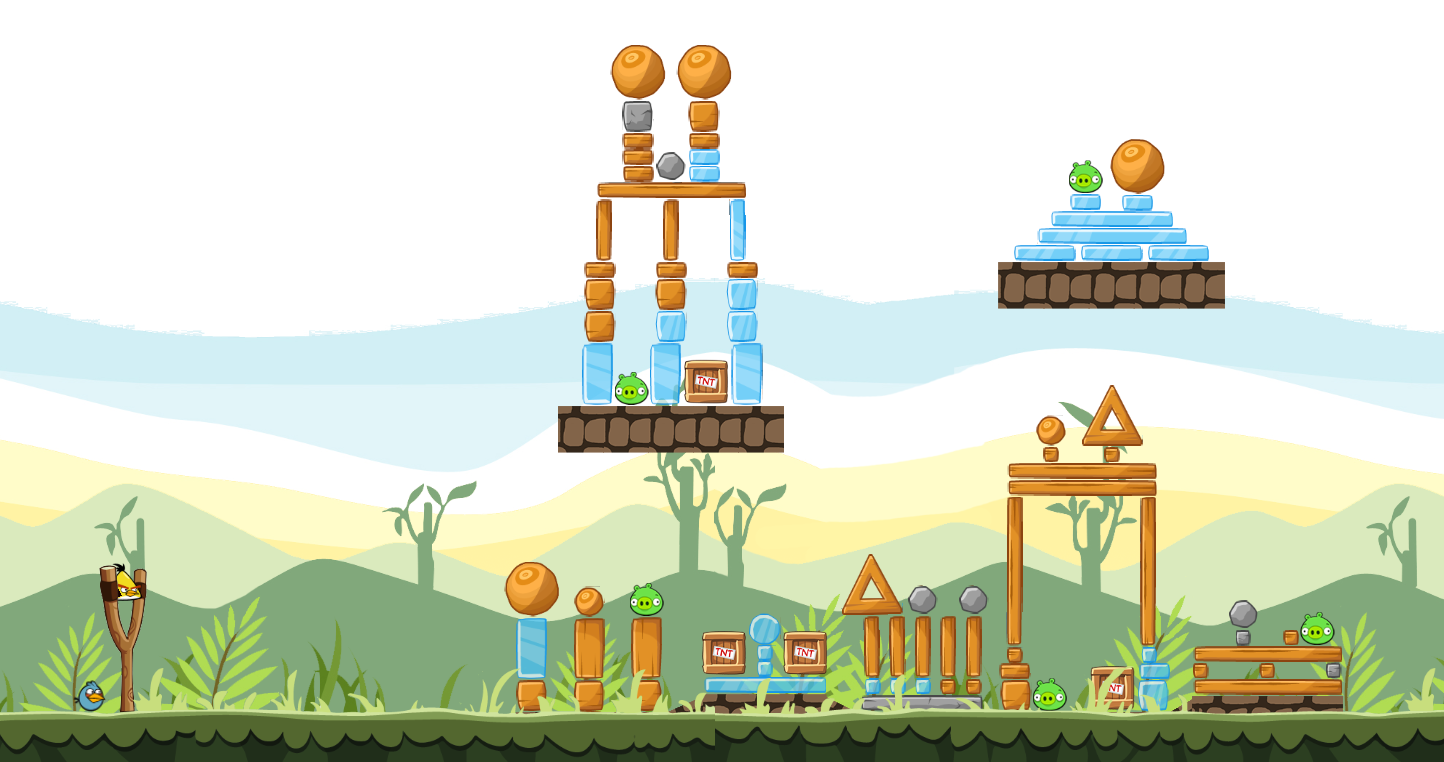}
  \caption{Generated level with a high $Fitness_m$ value.}
\end{figure}

\section{Discussion}
Using an agent-based adaption method to adjust the difficulty of generated levels has many potential uses. Being able to generate personalised content for human players has previously been shown to increase user engagement and overall enjoyment in games \cite{emo1}, but what we are most interested in discussing here is how adaptive level generation can be used to help improve agent development. Angry Birds agents that attempt to use some form of reinforcement learning to solve unknown levels have become increasingly popular over the past few years at the annual AIBirds competition \cite{new88,mine1}, but have so far failed to demonstrate any of the exceptional performance this technique has exhibited for many other games. In fact, many of these agents often rank among the lowest performing Angry Birds agents currently available.
One of the main reasons for this poor performance is believed to be a lack of available levels for training purposes, something that PLG can help address \cite{drlpcg}. We believe that the adaptive generation method proposed in this paper can potentially be used to improve the performance of reinforcement learning agents better than simply using randomly generated levels, and that adaptive generation can also be used to evaluate and help identify weaknesses within non-learning agents as well.

Firstly, for generating levels with a fixed percentage solve-rate for a specific agent ($Fitness_p$). When training an agent it is often desirable to focus on levels that the agent can occasionally solve, while still leaving plenty of areas to improve upon. Levels that the agent currently performs very well on every time do not give much new information to learn, whilst levels that the agent can never solve also give little information for the opposite reason.
This issue is especially important in a game like Angry Birds as reward is only given to the agent when it solves a level, making any accumulated score from previous shots meaningless if the level is not also solved (i.e. delayed reward). Generating adapted levels that a learning agent can currently solve some of the time (e.g. using a $D_{target}$ value of 50\%) will likely help the agent improve quicker. Although this hypothesis is yet to be demonstrated, it seems to us like a reasonably intuitive idea.

Secondly, for generating levels that are relatively hard for a specific agent compared to other agents ($Fitness_m$). Similar to using the $Fitness_p$ function, training on levels where our reinforcement learning agent performs poorly compared to other agents, that perhaps even use different AI techniques, might help to improve learning efficiency. Using this approach has the advantage that it can create levels which emphasise the learning agent's most obvious weaknesses more than others, ensuring that the learning agent's more pressing limitations are attended to first (i.e. ensures that the learning agent is at least on an equal performance to other agents before attempting to improve beyond this). 
This approach could also potentially be used for non-learning agents, allowing us to identify flaws in our strategies that need improving the most (i.e. understand where other agents are outperforming us). Another use is for benchmarking multiple or new agents, where it is often desirable to test on a collection of levels the produce a large variation in prior agent performance \cite{minecont}. This could be achieved by generating a small subset of levels with a high $Fitness_m$ value for each previous agent, and then combining these subsets together to give our benchmark set for a new or improved agent.

\section{Conclusions and Future Work}
In this paper we have presented an adaptive level generator for Angry Birds, that uses agents to adjust the difficulty of the generated levels based on the player's performance. Levels are generated using a search-based approach, with several different adjustable parameters. A genetic algorithm and fitness function based on agent performance is then used to evolve the generator's parameters over multiple generations. Levels can be generated for specific player solve-rates, or that are especially hard for the current player relative to the performance of others. Several experiments were conducted that demonstrated the effectiveness of our adaptive generator for both these requirements on a variety of agents. This approach can be used to create personalised levels for human players, as well as improving the usefulness of generated levels for training and evaluating agents.

While the experiments presented in this paper demonstrate the effectiveness of our adaptive generator (at least when using agents as human surrogates), there are several areas that could be improved in the future. The most obvious improvement would be to increase the number of generator input parameters that can be adjusted, as well as testing our method on a greater number of agents. We could also attempt to integrate the Iratus Aves level generator with other Angry Birds generators, increasing the variety of levels that could be created. It would also be good to analyse our adaptive generator's ability to cope with learning agents, whose performance might improve or change over time. We also hope to be able to investigate our hypothesis that adaptive level generation can improve the generality and effectiveness of reinforcement learning agents even more so than regular level generation algorithms. The approaches presented in this paper can also be easily extended to other physics-based puzzle games with similar mechanics.

\small
\bibliographystyle{named}
\bibliography{aaai19}

\end{document}